\documentclass{article}

\usepackage[nonatbib,main,final]{neurips_2026}

\usepackage[utf8]{inputenc} 
\usepackage[T1]{fontenc}    
\usepackage{hyperref}       
\usepackage{url}            
\usepackage{booktabs}       
\usepackage{amsfonts}       
\usepackage{nicefrac}       
\usepackage{microtype}      
\usepackage{xcolor}         
\usepackage{multirow} 
\usepackage[table]{xcolor}
\usepackage{graphicx}

\usepackage{amsmath}
\usepackage{amssymb}
\usepackage{bm}

\definecolor{colorfirst}{rgb}{0.76, 0.87, 0.92}   
\definecolor{colorsecond}{rgb}{.866,.945, 0.831}  
\definecolor{colorthird}{rgb}{1, 0.98, 0.83}      
\definecolor{colorcite}{rgb}{0.212, 0.490, 0.741}

\newcommand{\cellfirst}{\cellcolor{colorfirst}}
\newcommand{\cellsecond}{\cellcolor{colorsecond}}
\newcommand{\cellthird}{\cellcolor{colorthird}}

\newcommand{\firsttext}{\colorbox{colorfirst}}
\newcommand{\secondtext}{\colorbox{colorsecond}}
\newcommand{\thirdtext}{\colorbox{colorthird}}

\newcommand{\grayrow}{\rowcolor[HTML]{EFEFEF}}

\title{BridgeMVS: Bridging Multi-View Stereo and Monodepth via Bidirectional Dynamic Fusion }

\author{%
  Jianfei Jiang\thanks{ Equal contribution.}
  \quad Haochen Yu\footnotemark[1]
  \quad Qiankun Liu\thanks{Corresponding authors.}
  \quad Rui Yang \\
  \quad \textbf{Hongyuan Liu}
  \quad \textbf{Xianglong Meng}
  \quad \textbf{Jiansheng Chen}
  \quad \textbf{Huimin Ma\footnotemark[2]} \\
  University of Science and Technology Beijing, China \\
  \texttt{\{jiangjf,haochen.yu,ruiyang,hongyuanliu,mengxianglong\}@xs.ustb.edu.cn} \\ 
  \texttt{\{liuqk3,jschen,mhmpub\}@ustb.edu.cn} \\
}

\begin{document}

\maketitle

\begin{abstract}

Multi-View Stereo (MVS) and Monocular Depth Estimation (MDE) provide complementary cues for dense 3D reconstruction. While MVS leverages explicit geometry for metric depth, it struggles in challenging regions where feature matching is unreliable. Conversely, MDE provides robust structural priors but typically yields scale-ambiguous results. Existing hybrid methods mainly exploit monocular cues in a unidirectional manner, leaving the mutual enhancement between MDE and MVS insufficiently explored. To address this limitation, we propose BridgeMVS, a unified cascaded multi-stage MVS framework that bridges monocular depth features and multi-view cost volume representations through Bidirectional Dynamic Fusion (BDF). At each stage, BDF performs dynamic mono-to-cost and cost-to-mono updates, where monocular structural features guide cost volume refinement for more robust MVS depth estimation, while multi-view geometric evidence recalibrates monocular representations to improve their metric-scale consistency. Furthermore, the fused monocular and multi-view representations are progressively propagated from coarse to fine stages and constrained by stage-wise relative-scale MDE supervision and metric-scale MVS supervision, enabling cross-branch interactions to benefit subsequent high-resolution depth estimation. In addition, BDF is designed as a plug-and-play module that can be conveniently integrated into existing cascaded MVS frameworks. Extensive experiments demonstrate that BridgeMVS achieves state-of-the-art performance on the DTU, Tanks-and-Temples, and ETH3D benchmarks,  while simultaneously boosting the accuracy of monocular depth estimation. Notably, BridgeMVS ranks first on both the Tanks-and-Temples Intermediate and Advanced benchmarks. The source code is available at \href{https://github.com/JianfeiJ/BridgeMVS}{https://github.com/JianfeiJ/BridgeMVS}.

\end{abstract}

\section{Introduction}
\label{Introduction}

Multi-View Stereo (MVS) is a fundamental task in computer vision that aims to reconstruct dense 3D geometry from a set of calibrated images, which plays a crucial role in various applications~\cite{unposed-to-3d, spar-gs, xyzcylinder,autodrive}. Traditional MVS methods~\cite{gipuma, acmp, acmmp, apdmvs, hpmmvs, sedmvs, dvpmvs++} rely on handcrafted photo-consistency metrics, which limits their robustness in real-world scenarios. Learning-based MVS methods~\cite{mvsnet,rmvsnet} replace handcrafted matching costs with learned feature representations and differentiable cost volumes constructed via the plane-sweep algorithm~\cite{plane-sweep}, achieving substantial progress.

Despite this progress, existing learning-based MVS methods still struggle in challenging regions, such as weakly textured areas and reflective surfaces. In these regions, reliable correspondences across views are difficult to establish, resulting in inaccurate matching costs and ambiguous cost volumes. Existing methods mainly address this issue from two perspectives. One line of work improves feature representation by introducing CNN variants~\cite{cdsmvsnet}, Transformers~\cite{transmvsnet, etmvsnet, mvsformer++}, and Mamba~\cite{s6, mvsmamba} for feature aggregation. Another line focuses on cost volume regularization, evolving from single-stage 3D CNNs and RNNs~\cite{itermvs, effimvs, di-mvs, di-mvs++, diffmvs} to cascade frameworks~\cite{casmvsnet, cvpmvsnet, ucsnet}, and further incorporating Transformers~\cite{attention, wtmvsnet,costformer, mvsformer++, rrt-mvs} to aggregate matching costs and suppress unreliable responses. Nevertheless, the cost volume is still fundamentally built upon multi-view feature matching, making MVS vulnerable when visual correspondences are unreliable.

\begin{figure*}[!t]
  \centering
  \includegraphics[width=1.0\textwidth]{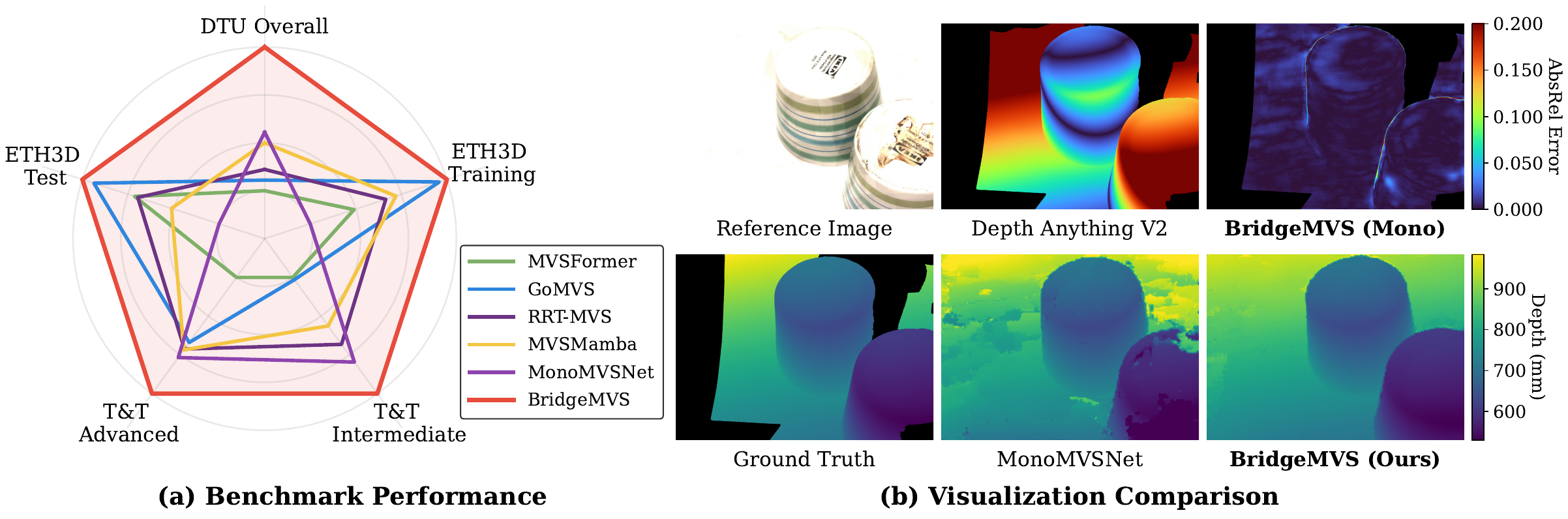}
  \caption{(a) Benchmark performance. Our method outperforms state-of-the-art methods~\cite{mvsformer, gomvs, rrt-mvs, mvsmamba, monomvsnet} on three standard MVS benchmarks~\cite{dtu, tanks, eth3d}. (b) Visualization comparison on Scan 48 from DTU~\cite{dtu}. Benefiting from bidirectional dynamic fusion, BridgeMVS improves MVS depth estimation~\cite{monomvsnet} in challenging regions and enhances the accuracy of monocular depth estimation~\cite{dav2} under multi-view geometric constraints. Absolute relative error maps are visualized after aligning the relative depth with GT using the ROE solver~\cite{moge}.}
  \label{fig:radar}
\end{figure*}

Recent Monocular Depth Estimation (MDE) models~\cite{dav1, dav2} provide promising complementary cues for MVS by encoding rich structural priors from large-scale training. These priors are naturally complementary to MVS: MDE offers strong structural awareness but predicts scale-ambiguous relative depth, while MVS provides explicit multi-view geometric constraints and metric-scale depth. Recent MonoMVSNet~\cite{monomvsnet} explores this complementarity by introducing monocular priors into the MVS pipeline. However, it mainly follows a unidirectional paradigm, where MDE only serves as external guidance for MVS. Such one-way interaction limits the use of monocular priors and prevents MDE from benefiting from multi-view geometric evidence, making it difficult to jointly improve metric MVS depth and relative monocular depth, as shown in Fig.~\ref{fig:radar}.

To address these limitations, we propose BridgeMVS, a unified cascaded MVS framework that bridges monocular depth features and multi-view cost volume representations through Bidirectional Dynamic Fusion (BDF). Instead of simply injecting monocular cues into the MVS branch, BDF performs bidirectional feature interaction at each cascade stage. In the mono-to-cost direction, monocular structural features are adapted to the cost volume space and used to guide cost volume refinement, helping the MVS branch distinguish ambiguous or redundant matching responses. In the cost-to-mono direction, geometry-aware information is aggregated from the regularized cost volume and fed back to the MDE branch, recalibrating intermediate monocular representations with explicit multi-view evidence. Through this bidirectional update, BridgeMVS jointly improves multi-view reconstruction and monocular depth estimation, leading to superior benchmark performance and more accurate depth predictions, as illustrated in Fig.~\ref{fig:radar}.

Furthermore, BridgeMVS exploits the inherent multi-scale representations of the DPT-based MDE branch and aligns them with the cascaded MVS structure. The fused monocular and multi-view representations are progressively propagated from coarse to fine stages, allowing early cross-branch interactions to benefit subsequent high-resolution depth estimation. Meanwhile, stage-wise relative-scale MDE supervision and metric-scale MVS supervision provide complementary structural and geometric constraints throughout the cascade. By preserving fused representations across stages, BridgeMVS enables monocular priors and multi-view geometric evidence to be gradually refined. Since BDF operates only on intermediate monocular features and cost volume representations, it can be used as a plug-and-play module for existing cascaded MVS frameworks.

The main contributions of this work are summarized as follows:

\begin{itemize}
\item We propose a plug-and-play Bidirectional Dynamic Fusion (BDF) module that performs mono-to-cost and cost-to-mono updates between monocular depth features and multi-view cost volumes, and can be conveniently integrated into existing cascaded MVS frameworks.

\item We propagate fused monocular and multi-view representations across cascaded stages from coarse to fine, and introduce stage-wise relative-scale and metric-scale supervision to enable progressive multi-scale refinement under both structural and geometric constraints.

\item Extensive experiments show that BridgeMVS achieves state-of-the-art performance on multiple MVS benchmarks, while also substantially improving monocular depth estimation.
\end{itemize}

\section{Related Work}

\subsection{Learning-based MVS}

Multi-View Stereo (MVS) reconstructs 3D geometry from calibrated multi-view images. MVSNet~\cite{mvsnet} firstly introduces a plane-sweep cost volume with 3D CNN-based regularization, establishing the standard learning-based MVS pipeline of feature extraction, cost volume construction, cost regularization, and depth prediction. Recent methods mainly improve MVS from two aspects. The first focuses on feature representation. Transformer-based methods~\cite{transmvsnet, etmvsnet, mvsformer, mvsformer++} enhance intra-view and inter-view feature aggregation with attention mechanisms, while MVSMamba~\cite{mvsmamba} introduces state-space modeling for efficient global feature aggregation. The second focuses on cost volume regularization. WT-MVSNet~\cite{wtmvsnet}, CostFormer~\cite{costformer}, and RRT-MVS~\cite{rrt-mvs} adopt window attention~\cite{swin} to aggregate matching costs, while MVSFormer++~\cite{mvsformer++} uses flash attention~\cite{flashattention} for cost regularization. GoMVS~\cite{gomvs} further incorporates normal priors to address local geometric inconsistency. Despite these advances, the cost volume is still fundamentally constrained to be constructed from multi-view feature matching, making MVS unreliable in weakly textured, reflective, or repetitive regions.

\subsection{MVS with MDE}

Monocular Depth Estimation (MDE)~\cite{depthpro} predicts depth from a single image and provides rich contextual and structural priors. Early models such as MiDaS~\cite{midas} and DPT~\cite{dpt} improve cross-dataset generalization with robust dense prediction architectures. Recent monocular foundation models, including Depth Anything~\cite{dav1,dav2} and MoGe~\cite{moge,moge2}, further demonstrate strong zero-shot generalization and produce coherent depth structures in challenging regions. Monocular priors have recently been introduced into stereo matching~\cite{defom,monster,bridgedepth,foundationstereo,ggev} and MVS to improve robustness when correspondence estimation is ambiguous. Among stereo matching methods, MonSter~\cite{monster} performs bidirectional refinement in the disparity space through scale-shift alignment and recurrent updates, while BridgeDepth~\cite{bridgedepth} employs cross-attention for single-stage feature alignment after DPT fusion. For MVS, MVSAnywhere~\cite{mvsanywhere} proposes a cost volume patchifier module to fuse cost volume and monocular encoder feature, and leverages large-scale training for zero-shot MVS depth estimation. MonoMVSNet~\cite{monomvsnet} introduces monocular features and monocular depth priors through feature guidance, dynamic depth sampling, and relative-consistency supervision. Concurrent work SPE-MVS~\cite{spe-mvs} incorporates monocular depth priors through spatial position encoding to enhance reconstruction performance. 
These MVS methods demonstrate the effectiveness of monocular priors but primarily use them to guide MVS in a unidirectional manner. In contrast, our method establishes multi-stage bidirectional interaction between intermediate DPT features and regularized multi-view cost volumes, enabling mutual enhancement of both MVS and MDE.

\section{Method}
\label{Method}

\begin{figure*}[!t]
  \centering
  \includegraphics[width=1.0\textwidth]{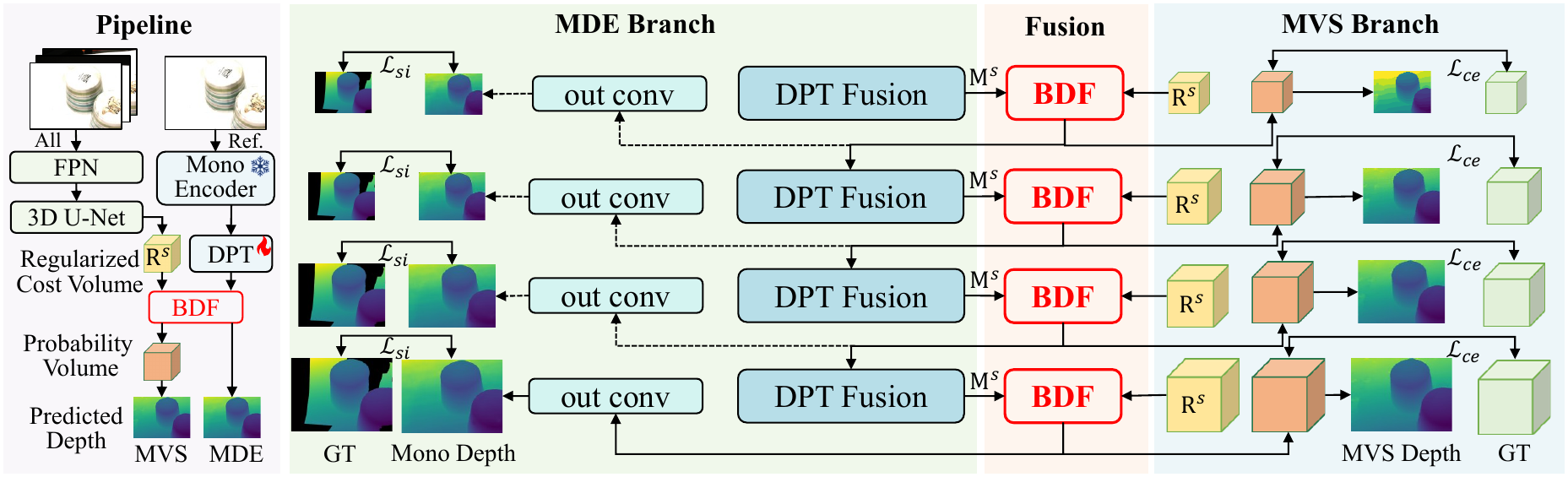}
  \caption{Overall architecture of BridgeMVS. The framework contains a cost volume-based MVS branch, a DPT-based MDE branch, and a Bidirectional Dynamic Fusion (BDF) module at each cascade stage. BDF bidirectionally fuses regularized cost volumes and monocular features, producing enhanced multi-view and monocular representations that are progressively propagated from coarse to fine stages under metric-scale MVS supervision and relative-scale MDE supervision.}
  \label{fig:overview}
\end{figure*}

\subsection{Overview}

The architecture of BridgeMVS is shown in Fig.~\ref{fig:overview}. Given $N$ calibrated images, including one reference image $I_0$ and $N-1$ source images $\{I_i\}_{i=1}^{N-1}$, BridgeMVS estimates both a metric MVS depth map and a relative monocular depth map for the reference image within a cascaded multi-stage MVS framework. The network consists of a cost volume-based MVS branch, a DPT-based MDE branch~\cite{dav2}, and a Bidirectional Dynamic Fusion (BDF) module (Sec.~\ref{sec:bdf}) inserted at each stage.

\textbf{MVS Branch}
follows the main pipeline of existing cascaded multi-scale MVS methods~\cite{casmvsnet, mvster} except for the inserted BDF module. Specifically, all images are fed into a Feature Pyramid Network (FPN)~\cite{fpn} to extract multi-scale image features. At stage $s$, each source-view feature is warped to the reference view to compute group-wise feature correlations~\cite{gwcnet, pvsnet} with the reference-view feature over the predefined depth hypotheses. Pixel-view weights~\cite{patchmatchnet} are then used to fuse all pairwise correlations into a single cost volume. Following common practice, a lightweight 3D U-Net~\cite{mvster} is applied for cost volume regularization. Instead of directly predicting the metric MVS depth from the regularized cost volume $\mathbf{R}^{s}$, BridgeMVS couples the regularized cost volume with the monocular feature (from the corresponding stage of the MDE branch) through BDF. The BDF-refined cost volume is converted into a probability volume by applying softmax along the depth dimension, and the metric MVS depth is obtained using a winner-take-all strategy. More details on cascaded multi-scale MVS can be found in~\cite{casmvsnet, mvster}.

\textbf{MDE Branch} follows the pipeline of the monocular depth foundation model (e.g., Depth Anything V2~\cite{dav2}) except for the BDF module. It consists of a DINOv2-based encoder~\cite{dinov2} and a DPT-based decoder~\cite{dpt}.
Different from the MVS branch, only the reference image is fed into the MDE branch.
The DPT decoder produces monocular features in a multi-stage manner, which naturally corresponds to the cascade multi-stage in the MVS branch. At stage $s$, the DPT monocular feature $\mathbf{M}^{s}$ is fused with the regularized cost volume $\mathbf{R}^{s}$ through BDF. The BDF-refined monocular feature  is decoded into a relative monocular depth map. Meanwhile, the BDF-refined monocular feature is propagated to the next stage in the DPT decoder, allowing multi-view geometric evidence injected at coarser stages to guide subsequent high-resolution monocular feature representations.

Through this design, BridgeMVS enables bidirectional interaction between multi-view and monocular representations at every cascade stage. The MVS branch benefits from monocular structural priors for robust cost volume refinement, while the MDE branch benefits from multi-view geometric constraints for improved relative consistency. The fused representations are progressively propagated from coarse to fine stages, producing both metric MVS depth and relative monocular depth.

\subsection{Bidirectional Dynamic Fusion}
\label{sec:bdf}

\begin{figure*}[!t]
  \centering
  \includegraphics[width=1.0\textwidth]{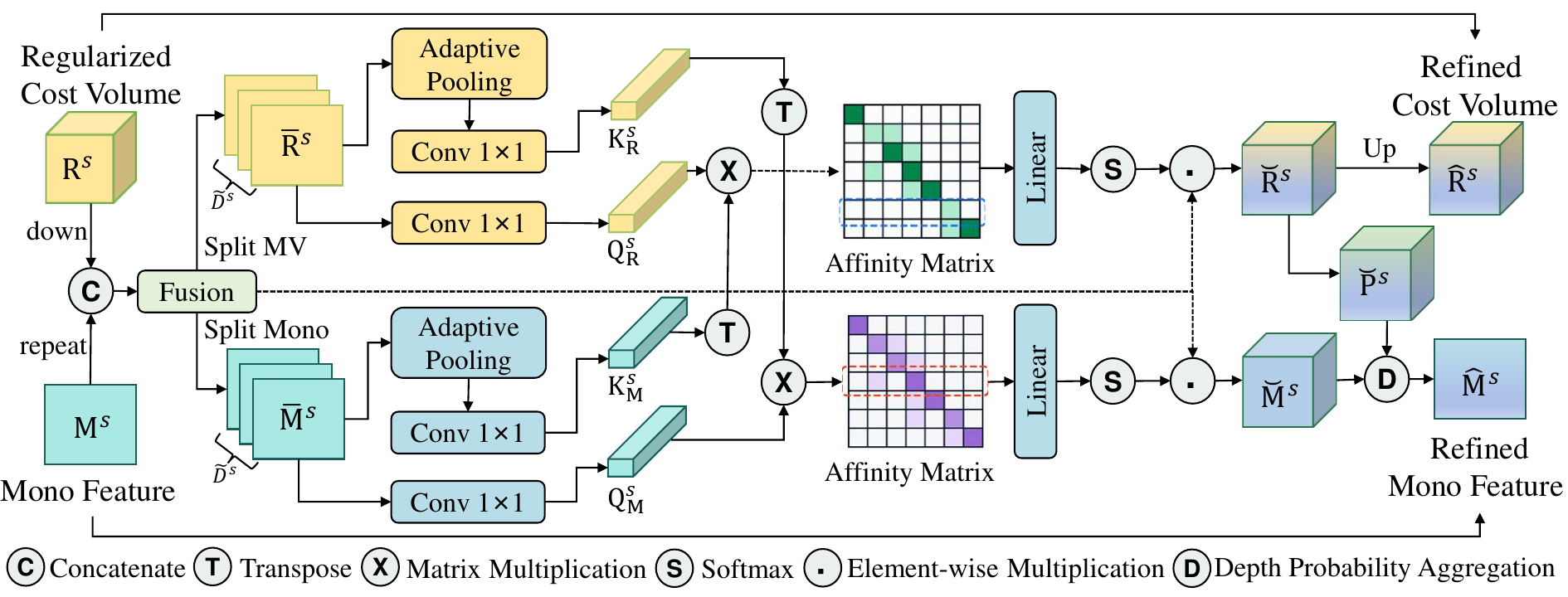}
  \caption{Overview of the proposed bidirectional dynamic fusion module. It first aligns and fuses monocular features and multi-view cost volumes, then performs bidirectional dynamic aggregation, followed by depth-guided feature fusion and residual update to produce refined multi-view cost volumes and monocular features.}
  \label{fig:bdf}
\end{figure*}

The proposed Bidirectional Dynamic Fusion (BDF) dynamically bridges the regularized cost volume and the DPT monocular feature through mono-to-cost aggregation and cost-to-mono depth probability fusion. As shown in Fig.~\ref{fig:bdf}, at stage $s$, BDF takes the regularized cost volume
$\mathbf{R}^{s} \in \mathbb{R}^{G^{s} \times D^{s} \times H^{s} \times W^{s}}$ (where $G^s$, $D^s$, and $H^s \times W^s$ are the number of groups, number of depth hypotheses and spatial resolution in the MVS branch)
and the corresponding DPT monocular feature
$\mathbf{M}^{s} \in \mathbb{R}^{C_{m}^{s} \times H_{m}^{s} \times W_{m}^{s}}$ (where $C_m^s$ and $H_m^s \times W_m^s$ are the channel dimension and spatial resolution in the MDE branch)
as inputs. Here, $\mathbf{R}^{s}$ encodes multi-view geometric evidence, while $\mathbf{M}^{s}$ provides strong monocular structural priors.

\textbf{Feature Alignment and Fusion.}
Since the two inputs have different forms and dimensions, we first align them into a shared 2D fusion space. For $\mathbf{R}^{s}$, it is downsampled by a 3D convolution layer and permuted along the first two dimensions to produce the aligned cost volume $\tilde{\mathbf{R}}^{s} \in \mathbb{R}^{\tilde{D}^{s} \times C_{v} \times \tilde{H}^{s} \times \tilde{W}^{s}}$, where $\tilde{D}^s$, $C_v$ and $\tilde{H}^{s} \times \tilde{W}^{s}$ are the number of depth hypotheses, the number of channels and the spatial resolution, respectively.
For $\mathbf{M}^{s}$, it is firstly bilinearly resized to $\tilde{H}^{s} \times \tilde{W}^{s}$ and projected to $C_{v}$ channels by a $1\times1$ 2D convolution, and then repeated along the depth dimension to produce the aligned monocular feature $\tilde{\mathbf{M}}^{s} \in \mathbb{R}^{\tilde{D}^{s} \times C_{v} \times \tilde{H}^{s} \times \tilde{W}^{s}}$.

In the following fusion and enhancement procedure, \emph{the depth dimension is independently processed} (similar to the batch dimension) since the monocular feature has no depth dimension in fact. 
The two aligned features are concatenated along the channel dimension and fed into a lightweight fusion module~\cite{unireplknet}. The fused representation is then split into an MVS representation $\bar{\mathbf{R}}_{s}$ and an MDE representation $\bar{\mathbf{M}}_{s}$, which are used for the subsequent bidirectional dynamic aggregation.

\textbf{Bidirectional Dynamic Aggregation.} 
The dynamic aggregation is inspired by context-mixing dynamic convolution~\cite{overlock}, which is implemented with the neighborhood attention~\cite{nat}. Differently, our bidirectional dynamic aggregation is mutual for both branches. As shown in Fig.~\ref{fig:bdf}, the mono-to-cost and cost-to-mono aggregation are equal in their implementation, but in an inverse direction.

Take the mono-to-cost aggregation for example, we first get the query $\mathbf{Q}_{\mathbf{R}}^{s} \in {\mathbb{R}^{\tilde{D}^{s} \times C_{v} \times \tilde{H}^{s} \tilde{W}^{s}}}$ from $\bar{\mathbf{R}}_{s}$ with a $1\times 1$ 2D convolution and a reshape operation. The key $\mathbf{K}_{\mathbf{M}}^{s} \in {\mathbb{R}^{\tilde{D}^{s} \times C_{v} \times \tilde{H}^{s} \tilde{W}^{s} / P^2}}$ is obtained from $\bar{\mathbf{M}}_{s}$ with an adaptive pooling, $1\times1$ convolution and a reshape operation, where $P$ is the pooling kernel size. Then, an affinity matrix $\mathbf{A}_{\mathbf{M} \rightarrow \mathbf{R}} \in \mathbb{R}^{\tilde{D}^{s} \times 1 \times \tilde{H}^{s} \tilde{W}^{s} \times \tilde{H}^{s} \tilde{W}^{s} / P^2}$ is obtained 
the inner product between $\mathbf{Q}_{\mathbf{R}}^{s}$ and $\mathbf{K}_{\mathbf{M}}^{s}$ along the channel dimension. 
by $\mathbf{A}_{\mathbf{M} \rightarrow \mathbf{R}} = (\mathbf{Q}_{\mathbf{R}}^{s})^{\top} \mathbf{K}_{\mathbf{M}}^{s}$.
The neighborhood attention matrix $\tilde{\mathbf{A}}_{\mathbf{M} \rightarrow \mathbf{R}} \in \mathbb{R}^{\tilde{D}^{s} \times 1 \times \tilde{H}^{s} \tilde{W}^{s} \times K^2}$ is finally produced by feeding $\mathbf{A}_{\mathbf{M} \rightarrow \mathbf{R}}$ to a linear layer (followed by a softmax function along the last dimension), where $K$ is the window size for neighborhood attention.
The aggregated cost volume $\check{\mathbf{R}}^{s} \in \mathbb{R}^{\tilde{D}^{s} \times C_{v} \times \tilde{H}^{s} \times \tilde{W}^{s}}$ is obtained by $\check{\mathbf{R}}^{s} = \text{NA}(\bar{\mathbf{R}}^{s}, \tilde{\mathbf{A}}_{\mathbf{M} \rightarrow \mathbf{R}})$, where $\text{NA}(\cdot, \cdot)$ is neighborhood attention proposed in~\cite{nat}.

Similarly, the aggregated monocular feature $\check{\mathbf{M}}^{s} \in \mathbb{R}^{\tilde{D}^{s} \times C_{v} \times \tilde{H}^{s} \times \tilde{W}^{s}}$ in the cost-to-mono direction can be obtained in a similar way. But differently, the query and key for affinity matrix computation are obtained from $\bar{\mathbf{M}}_{s}$ and $\bar{\mathbf{R}}_{s}$, respectively.

Finally, both 
$\check{\mathbf{R}}^{s}$ and $\check{\mathbf{M}}^{s}$ are permuted to the shape of $C_{v} \times \tilde{D}^{s} \times \tilde{H}^{s} \times \tilde{W}^{s}$, which are denoted as $\breve{\mathbf{R}}^{s}$ and $\breve{\mathbf{M}}^{s}$, respectively. In the mono-to-cost direction, monocular structural priors dynamically guide the aggregation of the MVS representation. In the cost-to-mono direction, multi-view geometric evidence dynamically updates the MDE representation. The feature representations are mutually enhanced by each other. 

\paragraph{Depth Probability Fusion and Residual Update.}
Since the MDE branch does not require the depth dimension, the depth dimension in $\breve{\mathbf{M}}^{s}$ needs to be reduced.
To do this, a depth probability volume $\breve{\mathbf{P}}^{s} \in \mathbb{R}^{1 \times \tilde{D}^{s} \times \tilde{H}^{s} \times \tilde{W}^{s}}$ is obtained by mapping $\breve{\mathbf{R}}^{s}$ with a lightweight probability projection layer (followed by a softmax activation along the depth dimension). The reduced aggregated monocular feature $\breve{\mathbf{M}}^{s}_{r} \in \mathbb{R}^{C_v \times \tilde{H}^{s} \times \tilde{W}^{s}}$  is obtained as the element-wise inner product between $\breve{\mathbf{P}}^{s}$ and $\breve{\mathbf{M}}^{s}$ along the depth dimension.

Both the $\breve{\mathbf{R}}^{s}$ and $\breve{\mathbf{M}}^{s}_{r}$ are mapped to the original cost volume space and monocular feature space with convolution layers, and used as the residual update, which are added to $\mathbf{R}^{s}$ and $\mathbf{M}^{s}$ respectively, producing the enhanced cost volume $\hat{\mathbf{R}}^{s} \in \mathbb{R}^{G^{s} \times D^{s} \times H^{s} \times W^{s}}$ and monocular feature $\hat{\mathbf{M}}^{s} \in \mathbb{R}^{C_{m}^{s} \times H_{m}^{s} \times W_{m}^{s}}$. The enhanced cost volume $\hat{\mathbf{R}}^{s}$ and monocular feature $\hat{\mathbf{M}}^{s}$ are used for the depth prediction in stage $s$, but also propagated to the following stage in their corresponding branches.

\subsection{Training Objective}

BridgeMVS is supervised at each cascade stage by both the MVS branch and the MDE branch. The MVS branch follows common cascaded multi-stage MVS methods and adopts the cross-entropy loss over the predicted depth probability volume, which provides metric-scale supervision for MVS depth estimation. The MDE branch applies the scale-invariant loss~\cite{midas} to the relative monocular depth prediction, which constrains monocular depth in a scale-invariant manner and preserves its structural prior.
Specifically, let $\hat{\mathbf{P}}^{s}, \mathbf{P}_{gt}^{s} \in \mathbb{R}^{D^{s} \times H^{s} \times W^{s}}$ be the predicted probability volume and ground-truth one-hot probability volume for stage $s$ in the MVS branch, and $\hat{\mathbf{D}}^{s}, \mathbf{D}^{s}_{gt} \in  \mathbb{R}^{H^{s}_{m} \times W^{s}_{m}}$ be the predicted depth map and ground-truth depth map for stage $s$ in the MDE branch, the overall training objective is defined as:
\begin{equation}
\mathcal{L}
=
\sum_{s}
\left[
\lambda_{mvs}^{s}
\mathcal{L}_{ce}
\left(
\hat{\mathbf{P}}^{s},
\mathbf{P}_{gt}^{s}
\right)
+
\lambda_{mde}^{s}
\mathcal{L}_{si}
\left(
\hat{\mathbf{D}}^{s},
\mathbf{D}_{gt}^{s}
\right)
\right],
\end{equation}
where $\lambda_{mvs}^{s}$ and $\lambda_{mde}^{s}$ are stage-wise loss weights. The metric-scale MVS supervision and relative-scale MDE supervision jointly constrain the coarse-to-fine propagation of fused representations.

\section{Experiments}
\label{Experiments}

\begin{figure}[!t]
  \centering
  \includegraphics[width=1.0\linewidth]{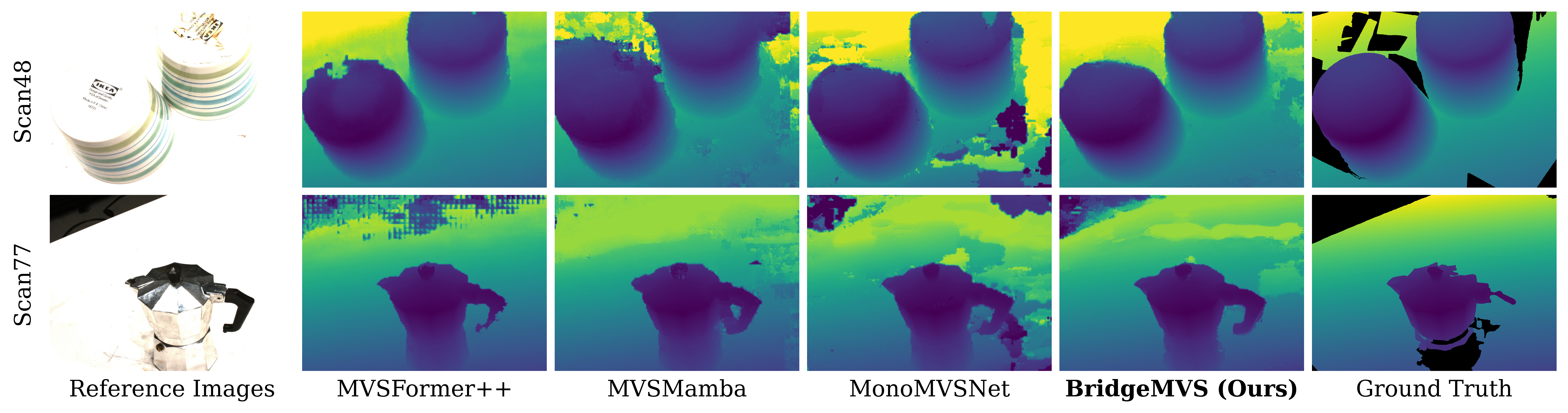}
  \caption{Qualitative comparison of depth maps in most challenging scan48 and scan 77 on DTU. Our method predicted more accurate depth maps in weakly textured and reflection areas.}
  \label{fig:dtu_depth}
\end{figure}

\subsection{Datasets and Benchmarks}

DTU~\cite{dtu} is an indoor MVS dataset captured under controlled laboratory conditions. BlendedMVS~\cite{blendedmvs} is a large-scale synthetic dataset containing both indoor and outdoor scenes. Tanks-and-Temples (T\&T)~\cite{tanks} is a real-world MVS benchmark with complex geometry and diverse acquisition conditions. It includes 8 scenes in the intermediate set and 6 scenes in the advanced set. ETH3D~\cite{eth3d} benchmark contains high-resolution indoor and outdoor scenes with large viewpoint variations, and is divided into training and test sets. RobustMVD~\cite{rmvd} is a multi-view depth estimation benchmark comprising five datasets: KITTI~\cite{kitti}, ScanNet~\cite{scannet}, ETH3D~\cite{eth3d}, DTU~\cite{dtu}, and T\&T~\cite{tanks}.

\subsection{Implementation Details}
\label{Implementation Details}

BridgeMVS is implemented in PyTorch~\cite{pytorch} and optimized using Adam~\cite{adam}. Following common practice, we first train and evaluate BridgeMVS on DTU~\cite{dtu}. For DTU training, we use 5 input views at a resolution of 512$\times$640 for 15 epochs, with an initial learning rate of 1e-3. To improve generalization, we further fine-tune the model on a mixture of DTU and BlendedMVS~\cite{blendedmvs}, using 11 input views at a resolution of 512$\times$640 for an additional 15 epochs, with an initial learning rate of 4e-4. In addition, following~\cite{mvster,mvsformer,mvsformer++,mvsmamba}, we conduct high-resolution training on DTU using 5 input views at a resolution of 1024$\times$1280 for 10 epochs, with an initial learning rate of 1e-3 (the model is denoted as BridgeMVS*). We use the ViT-B version of Depth Anything V2~\cite{dav2} as the MDE branch, freezing the encoder and fine-tuning the DPT decoder with a learning rate 0.05$\times$ that of the rest of the network. BridgeMVS uses four coarse-to-fine stages with 32-16-8-4 depth hypotheses and 4-4-4-4 grouped feature correlations. 
We using the dynamic fusion strategy~\cite{d2hcrmvsnet} to obtain the final reconstructed point clouds.

\subsection{Benchmark Performance}

\begin{table}[!t]
  \centering
  \small
  \caption{Quantitative results of reconstructed point clouds on the DTU~\cite{dtu} evaluation set with distance metric [mm]. The \firsttext{best}, \secondtext{second-best}, and \thirdtext{third-best} results are highlighted with colors, respectively. Methods marked with * use high-resolution images for training.}
    \begin{tabular}{lccc}
    \toprule
    Methods & \textbf{Overall} $\downarrow$ & Accuracy $\downarrow$ & Completeness $\downarrow$ \\
    \midrule
    CasMVSNet~\cite{casmvsnet} & 0.355 & 0.325 & 0.385 \\
    UniMVSNet~\cite{unimvsnet} & 0.315 & 0.352 & 0.278 \\
    MVSTER*~\cite{mvster} & 0.313 & 0.350 & 0.276 \\
    TransMVSNet~\cite{transmvsnet} & 0.305 & 0.321 & 0.289 \\
    DMVSNet~\cite{dmvsnet} & 0.305 & 0.338 & 0.272 \\
    GeoMVSNet~\cite{geomvsnet} & 0.295 & 0.331 & 0.259 \\
    ET-MVSNet~\cite{etmvsnet} & 0.291 & 0.329 & 0.253 \\
    MVSFormer*~\cite{mvsformer} & 0.289 & 0.327 & 0.251 \\
    GoMVS~\cite{gomvs} & 0.287 & 0.347 & \cellsecond 0.227 \\
    RRT-MVS~\cite{rrt-mvs} & 0.285 & 0.309 & 0.261 \\
    MVSFormer++*~\cite{mvsformer++} & 0.281 & 0.309 & 0.252 \\
    MVSMamba*~\cite{mvsmamba} & 0.280 & \cellthird 0.308 & 0.252 \\
    MonoMVSNet~\cite{monomvsnet} & \cellthird 0.278 & 0.313 & 0.243 \\
    MVSAnywhere~\cite{mvsanywhere} & 0.735 & 0.845 & 0.625 \\
    \midrule
    \textbf{BridgeMVS (Ours)} & \cellsecond 0.268 & \cellfirst 0.297 & \cellthird 0.239 \\
    \textbf{BridgeMVS* (Ours)} & \cellfirst 0.262 & \cellsecond 0.303 & \cellfirst 0.221 \\
    \bottomrule
    \end{tabular}
  \label{tab:dtu}
\end{table}

\begin{table}[!t]
\centering
\caption{
Quantitative point cloud results on the T\&T~\cite{tanks} benchmark with F-score (\%).
The \firsttext{best}, \secondtext{second-best}, and \thirdtext{third-best} results are marked with colors.}
\resizebox{\textwidth}{!}{
\begin{tabular}{l*{16}{c}}
\toprule
\multirow{2}{*}{Methods} & \multicolumn{9}{c}{Intermediate set $\uparrow$} & \multicolumn{7}{c}{Advanced set $\uparrow$} \\
\cmidrule(lr){2-10}
\cmidrule(lr){11-17}
& \textbf{Mean} & Fam. & Fra. & Hor. & L.H. & M60 & Pan. & P.G. & Tra.
& \textbf{Mean} & Aud. & Bal. & Cou. & Mus. & Pal. & Tem. \\
\midrule

CasMVSNet~\cite{casmvsnet} 
& 56.84 & 76.37 & 58.45 & 46.26 & 55.81 & 56.11 & 54.06 & 58.18 & 49.51
& 31.12 & 19.81 & 38.46 & 29.10 & 43.87 & 27.36 & 28.11 \\

TransMVSNet~\cite{transmvsnet} 
& 63.52 & 80.92 & 65.83 & 56.94 & 62.54 & 63.06 & 60.00 & 60.20 & 58.67
& 37.00 & 24.84 & 44.59 & 34.77 & 46.49 & 34.69 & 36.62 \\

MVSTER~\cite{mvster} 
& 60.92 & 80.21 & 63.51 & 52.30 & 61.38 & 61.47 & 58.16 & 58.98 & 51.38
& 37.53 & 26.68 & 42.14 & 35.65 & 49.37 & 32.16 & 39.19 \\

UniMVSNet~\cite{unimvsnet} 
& 64.36 & 81.20 & 66.43 & 53.11 & 63.46 & \cellsecond 66.09 & 64.84 & 62.23 & 57.53
& 38.96 & 28.33 & 44.36 & 39.74 & 52.89 & 33.80 & 34.63 \\

ET-MVSNet~\cite{etmvsnet} 
& 65.49 & 81.65 & 68.79 & 59.46 & 65.72 & 64.22 & 64.03 & 61.23 & 58.79
& 40.41 & 28.86 & 45.18 & 38.66 & 51.10 & 35.39 & 43.23 \\

MVSFormer~\cite{mvsformer} 
& 66.38 & 82.06 & 69.34 & 60.49 & 68.61 & 65.67 & 64.08 & 61.23 & 59.53
& 40.87 & 28.22 & 46.75 & 39.30 & 52.88 & 35.16 & 42.95 \\

DMVSNet~\cite{dmvsnet} 
& 64.66 & 81.27 & 67.54 & 59.10 & 63.12 & 64.64 & 64.80 & 59.83 & 56.97
& 41.17 & 30.08 & 46.10 & 40.65 & 53.53 & 35.08 & 41.60 \\

GeoMVSNet~\cite{geomvsnet} 
& 65.89 & 81.64 & 67.53 & 55.78 & 68.02 & 65.49 & \cellthird 67.19 & 63.27 & 58.22
& 41.52 & 30.23 & 46.54 & 39.98 & 53.05 & 35.98 & 43.34 \\

MVSFormer++~\cite{mvsformer++} 
& 67.18 & \cellsecond 82.69 & 69.44 & \cellthird 64.24 & 69.16 & 64.13 & 66.43 & 61.19 & 60.12
& 41.60 & 29.93 & 45.69 & 39.46 & 53.58 & 35.56 & 45.39 \\

GoMVS~\cite{gomvs} 
& 66.44 & \cellthird 82.68 & 69.23 & \cellfirst 69.19 & 63.56 & 65.13 & 62.10 & 58.81 & \cellsecond 60.80
& 43.07 & \cellfirst 35.52 & \cellthird 47.15 & \cellthird 42.52 & 52.08 & 36.34 & 44.82 \\

RRT-MVS~\cite{rrt-mvs} 
& \cellthird 68.16 & 82.54 & 72.31 & 61.44 & \cellthird 69.89 & 65.32 & \cellfirst 68.88 & 64.45 & 60.48
& 43.29 & \cellthird 30.95 & 46.42 & 41.13 & \cellsecond 55.46 & \cellfirst 37.63 & \cellsecond 48.12 \\

MVSMamba~\cite{mvsmamba} 
& 67.67 & 82.47 & \cellsecond 72.90 & 58.55 & 69.63 & 65.34 & 66.88 & \cellfirst 65.60 & 59.98
& \cellthird 43.32 & \cellthird 30.95 & \cellfirst 49.61 & 41.04 & 54.92 & 36.72 & 46.67 \\

MonoMVSNet~\cite{monomvsnet} 
& \cellsecond 68.63 & 82.38 & \cellthird 72.89 & 62.80 & \cellfirst 70.49 & \cellthird 65.79 & \cellsecond 68.54 & \cellsecond 65.54 & \cellthird 60.59
& \cellsecond 43.58 & 30.33 & 46.76 & \cellfirst 42.90 & \cellfirst 56.31 & \cellsecond 37.28 & \cellthird 47.88 \\

\midrule

\textbf{BridgeMVS (Ours)} 
& \cellfirst 69.47 & \cellfirst 83.25 & \cellfirst 73.81 & \cellsecond 66.80 & \cellsecond 70.32 & \cellfirst 66.30 & \cellfirst 68.88 & \cellthird 64.57 & \cellfirst 61.82
& \cellfirst 44.80 & \cellsecond 34.25 & \cellsecond 49.06 & \cellsecond 42.86 & \cellthird 55.35 & \cellthird 36.83 & \cellfirst 50.44 \\

\bottomrule
\end{tabular}
}
\label{tab:tnt}
\end{table}

\paragraph{Evaluation on DTU.} We evaluate BridgeMVS and BridgeMVS* on DTU~\cite{dtu} evaluation set, using the standard 5 input views at resolutions of 832$\times$1152 and 1152$\times$1600, respectively. We use the official evaluation script to report accuracy, completeness, and their average overall error. As shown in Tab.~\ref{tab:dtu}, both BridgeMVS and BridgeMVS* outperform previous state-of-the-art methods by a large margin, achieving overall errors of $\mathbf{0.268}$ and $\mathbf{0.262}$, respectively. Qualitative comparisons are shown in Fig.~\ref{fig:dtu_depth}, with more results provided in Fig.~\ref{fig:bridgemvs_dtu_depth_all} of Appendix~\ref{sec:more_vis}. Compared with MonoMVSNet~\cite{monomvsnet}, which also exploits monocular priors, BridgeMVS improves all metrics and predicts more accurate depth maps, while maintaining nearly the same GPU memory consumption and running time, as shown in Tab.~\ref{tab:vit}. These results demonstrate that bidirectional interaction between monocular features and multi-view cost volumes is more effective than simple unidirectional monocular guidance.

\paragraph{Evaluation on T\&T.} To assess the generalization capability, we evaluate BridgeMVS on T\&T~\cite{tanks} benchmark with 21 input views. Tab.~\ref{tab:tnt} presents the quantitative results. BridgeMVS surpasses all other state-of-the-art methods with mean F-score of \textbf{69.47} and \textbf{44.80} on both Intermediate and Advanced sets, respectively. Additional qualitative results are shown in Fig.~\ref{fig:bridgemvs_tnt} of Appendix~\ref{sec:more_vis}.

\paragraph{Evaluation on ETH3D.}
To further assess the generalization ability, we evaluate BridgeMVS on the challenging ETH3D~\cite{eth3d} high-resolution benchmark using 11 input views at a resolution of $1600 \times 2432$. As shown in Tab.~\ref{tab:eth3d}, BridgeMVS achieves the best $F_1$-score on both the training and test sets. ETH3D contains high-resolution scenes with large baselines and strong viewpoint variations, posing significant challenges for learning-based MVS methods. Since recent state-of-the-art methods on T\&T, such as RRT-MVS~\cite{rrt-mvs}, MVSMamba~\cite{mvsmamba} and MonoMVSNet~\cite{monomvsnet}, have not reported results on ETH3D, we evaluate them using their open-source code and pretrained weights. The clear performance gains demonstrate the strong generalization ability of BridgeMVS.

\begin{table*}[!t]
\centering
\small
\caption{Quantitative point cloud results on the ETH3D~\cite{eth3d} high-res benchmark with $F_1$-score [\%] at the evaluation threshold of 2 cm. The \firsttext{best}, \secondtext{second-best}, and \thirdtext{third-best} results are marked with colors, respectively. Methods marked with $^\dag$ are reproduced by using their open-source code.}
\begin{tabular}{lcccccc}
\toprule
\multirow{2}{*}{Methods} & \multicolumn{3}{c}{Training Set} & \multicolumn{3}{c}{Test Set} \\
\cmidrule(lr){2-4} \cmidrule(lr){5-7}
& \textbf{\boldmath$F_1$-score} $\uparrow$ & Precision $\uparrow$ & Recall $\uparrow$ 
& \textbf{\boldmath$F_1$-score} $\uparrow$ & Precision $\uparrow$ & Recall $\uparrow$ \\
\midrule

MonoMVSNet$^\dag$~\cite{monomvsnet} 
& 70.50 & 72.96 & 68.80 
& 76.47 & 76.66 & 76.99 \\

MVSTER~\cite{mvster} 
& 72.06 & 68.08 & \cellthird 76.92 
& 79.01 & 77.09 & 82.47 \\

MVSMamba$^\dag$~\cite{mvsmamba} 
& 76.25 & 77.85 & 75.14 
& 80.06 & 80.30 & 81.55 \\

RRT-MVS$^\dag$~\cite{rrt-mvs} 
& 75.57 & 76.82 & 75.00 
& 82.60 & 82.97 & 82.61 \\

MVSFormer~\cite{mvsformer} 
& 73.44 & 73.62 & 74.64 
& 82.85 & 82.23 & 83.75 \\

MVSFormer++~\cite{mvsformer++} 
& - & - & - 
& 82.99 & 81.88 & 83.88 \\

EPP-MVSNet~\cite{eppmvsnet} 
& 74.00 & \cellsecond 82.76 & 67.58 
& 83.40 & \cellthird 85.47 & 81.79 \\

VisMVSNet~\cite{vismvsnet} 
& 72.77 & \cellfirst 83.32 & 65.53 
& 83.46 & \cellfirst 86.86 & 80.92 \\

Effi-MVS+~\cite{effimvs++} 
& 75.28 & \cellthird 82.33 & 70.42 
& 83.62 & 83.90 & 83.81 \\

CasDiffMVS~\cite{diffmvs}
& \cellthird 76.75 & 79.93 & 75.20 
& \cellthird 85.11 & 85.21 & \cellthird 85.37 \\

GoMVS~\cite{gomvs} 
& \cellsecond 79.16 & 81.22 & \cellsecond 77.65 
& \cellsecond 85.91 & \cellsecond 85.50 & \cellsecond 86.85 \\

\midrule

\textbf{BridgeMVS} 
& \cellfirst 79.68 & 79.44 & \cellfirst 80.40 
& \cellfirst 86.78 & 84.80 & \cellfirst 89.08 \\

\bottomrule
\end{tabular}
\label{tab:eth3d}
\end{table*}

\begin{table*}[!t]
\centering
\small
\setlength{\tabcolsep}{4pt}
\caption{Quantitative depth estimation results on the RobustMVD~\cite{rmvd} benchmark.
$\dagger$: relative depth are alignment with GT using the ROE~\cite{moge} solver.
$\ddagger$: trained on eight datasets.}
\begin{tabular}{lcccccccccccc}
\toprule
\multirow{2}{*}{Methods}
& \multicolumn{2}{c}{DTU}
& \multicolumn{2}{c}{T\&T}
& \multicolumn{2}{c}{ETH3D}
& \multicolumn{2}{c}{KITTI}
& \multicolumn{2}{c}{ScanNet}
& \multicolumn{2}{c}{Average} \\
\cmidrule(lr){2-3}
\cmidrule(lr){4-5}
\cmidrule(lr){6-7}
\cmidrule(lr){8-9}
\cmidrule(lr){10-11}
\cmidrule(lr){12-13}
& rel $\downarrow$ & $\tau$ $\uparrow$
& rel $\downarrow$ & $\tau$ $\uparrow$
& rel $\downarrow$ & $\tau$ $\uparrow$
& rel $\downarrow$ & $\tau$ $\uparrow$
& rel $\downarrow$ & $\tau$ $\uparrow$
& rel $\downarrow$ & $\tau$ $\uparrow$ \\
\midrule
Depth Anything V2$^\dagger$~\cite{dav2}
& 5.9 & 53.2
& 7.0 & 50.5
& 7.9 & 42.3
& 7.1 & 41.2
& 12.7 & 24.3
& 8.1 & 42.3 \\
BridgeMVS (Mono)$^\dagger$
& 0.9 & 95.5
& 3.0 & 77.9
& 4.9 & 58.9
& 6.0 & 45.6
& 5.5 & 49.2
& 4.1 & 65.4 \\
\midrule
MonoMVSNet~\cite{monomvsnet}
& 2.8 & 81.1
& 6.5 & 81.4
& 7.5 & 50.0
& 5.1 & 58.2
& 9.7 & 31.8
& 6.3 & 60.5 \\
MVSAnywhere~\cite{mvsanywhere}
& 1.4 & 90.7
& 2.7 & 80.1
& \textbf{3.6} & \textbf{61.5}
& 4.1 & 57.8
& \textbf{4.8} & \textbf{53.7}
& \textbf{3.3} & 68.8 \\
BridgeMVS (Ours)
& \textbf{0.9} & \textbf{94.4}
& \textbf{2.5} & \textbf{89.8}
& 4.9 & 57.3
& \textbf{3.7} & \textbf{66.5}
& 6.8 & 40.3
& 3.8 & \textbf{69.7} \\
\midrule
\textcolor{gray}{MVSAnywhere (full)$^\ddagger$}
& \textcolor{gray}{1.3} & \textcolor{gray}{95.0}
& \textcolor{gray}{2.1} & \textcolor{gray}{90.5}
& \textcolor{gray}{3.2} & \textcolor{gray}{68.0}
& \textcolor{gray}{3.2} & \textcolor{gray}{68.8}
& \textcolor{gray}{3.7} & \textcolor{gray}{62.9}
& \textcolor{gray}{2.7} & \textcolor{gray}{77.0} \\
\bottomrule
\end{tabular}
\label{tab:rmvd}
\end{table*}

\paragraph{Evaluation on RobustMVD.}
Tab.~\ref{tab:rmvd} evaluates depth estimation on RobustMVD~\cite{rmvd}. We report absolute relative error (rel) and inlier percentage $\tau$ with a threshold of $1.03$.
BridgeMVS (Mono) improves over the baseline Depth Anything V2~\cite{dav2}
across all five datasets, showing that the monocular branch
benefits from multi-view geometry transferred through BDF.
Under matched DTU+BlendedMVS training data, BridgeMVS outperforms
MonoMVSNet~\cite{monomvsnet} and surpasses MVSAnywhere~\cite{mvsanywhere}
on DTU, T\&T, and KITTI, with a higher average $\tau$.
MVSAnywhere performs better on ScanNet and ETH3D, where limited
camera motion or large viewpoint changes may weaken matching
evidence and favor its DPT decoder's use of monocular priors.
Although MVSAnywhere (full) achieves better average
depth metrics with training on eight datasets, BridgeMVS produces
better DTU point clouds (Tab.~\ref{tab:dtu}), demonstrating its
strength in multi-view reconstruction.
More qualitative results appear in Fig.~\ref{fig:bridgemvs_rmvd}
of Appendix~\ref{sec:more_vis}.

\begin{table}[!t]
    \centering
    \small
    \caption{Structure analysis of Bidirectional Dynamic Fusion (BDF). \textcolor[gray]{0.5}{Gray numbers} indicate results from the pre-trained monocular foundation model~\cite{dav2}.}
    \begin{tabular}{lccccccc}
    \toprule
    \multirow{2}{*}{Methods (ViT-B)} & \multicolumn{5}{c}{MVS Branch} & \multicolumn{2}{c}{MDE Branch} \\
    \cmidrule(lr){2-6} \cmidrule(lr){7-8}
    & Overall $\downarrow$ & Acc. $\downarrow$ & Comp. $\downarrow$ & MAE $\downarrow$ & RMSE $\downarrow$ & rel $\downarrow$ & $\tau$ $\uparrow$ \\
    \midrule
    (a) Base (w/o BDF) & 0.292 & 0.315 & 0.269 & 10.66 & 30.59 & \textcolor[gray]{0.5}{8.30} & \textcolor[gray]{0.5}{59.6} \\
    \midrule
    (b) Mono (frozen) $\rightarrow$ Cost & 0.279 & 0.312 & 0.246 & 7.61 & 23.11 & 2.77 & 71.4 \\
    (c) Mono (trainable) $\rightarrow$ Cost & 0.276 & 0.309 & 0.243 & 6.83 & 21.76 & 2.15 & 80.0 \\
    (d) Cost $\rightarrow$ Mono (trainable) & 0.289 & 0.313 & 0.265 & 9.04 & 26.27 & 0.95 & \textbf{94.7} \\
    \midrule
    (e) BDF (w/o pretrain) & 0.281 & 0.311 & 0.251 & 8.05 & 23.71 & 1.16 & 92.5 \\
    (f) BDF (w/o mono loss) & 0.273 & 0.307 & \textbf{0.239} & 7.36 & 22.56 & 8.09 & 30.7 \\
    \grayrow (g) BDF (Ours) & \textbf{0.268} & \textbf{0.297} & \textbf{0.239} & \textbf{6.56} & \textbf{20.71} & \textbf{0.92} & 93.6 \\
    \bottomrule
    \end{tabular}
    \label{tab:ablation}
\end{table}

\begin{table}[!t]
    \centering
    \small
    \setlength{\tabcolsep}{3pt}
    \caption{Compatibility study of Bidirectional Dynamic Fusion (BDF) as a plug-in.}
    \begin{tabular}{lccccccccc}
    \toprule
    \multirow{2}{*}{Methods (ViT-S)}
    & \multicolumn{5}{c}{MVS Branch}
    & \multicolumn{2}{c}{MDE Branch}
    & \multirow{2}{*}{Mem. (MB) $\downarrow$}
    & \multirow{2}{*}{Time (s) $\downarrow$} \\
    \cmidrule(lr){2-6} \cmidrule(lr){7-8}
    & Overall $\downarrow$
    & Acc. $\downarrow$
    & Comp. $\downarrow$
    & MAE $\downarrow$
    & RMSE $\downarrow$
    & rel $\downarrow$
    & $\tau$ $\uparrow$
    & & \\
    \midrule
    MVSMamba
    & 0.287 & 0.314 & 0.260 & 9.59 & 27.78
    & \textcolor[gray]{0.5}{8.30}
    & \textcolor[gray]{0.5}{59.6}
    & \textbf{2090} & \textbf{0.11} \\
    \grayrow + BDF
    & \textbf{0.275} & \textbf{0.291} & \textbf{0.259}
    & \textbf{7.43} & \textbf{22.84}
    & \textbf{0.93} & \textbf{93.6}
    & 2706 & 0.27 \\
    \midrule
    MonoMVSNet
    & 0.278 & 0.313 & 0.243 & 12.94 & 34.65
    & \textcolor[gray]{0.5}{8.30}
    & \textcolor[gray]{0.5}{59.6}
    & \textbf{2822} & \textbf{0.24} \\
    \grayrow + BDF
    & \textbf{0.270} & \textbf{0.304} & \textbf{0.236}
    & \textbf{9.43} & \textbf{24.91}
    & \textbf{1.19} & \textbf{91.2}
    & 3082 & 0.31 \\
    \bottomrule
    \end{tabular}
    \label{tab:plug}
\end{table}

\subsection{Ablation Study}
\label{sec:ablation}

We conduct ablation studies on the DTU~\cite{dtu} evaluation set using both point cloud and depth metrics. For MVS depth estimation, we report MAE and RMSE. For monocular depth estimation, we report rel and $\tau$. Unless otherwise specified, all ablation experiments use the model trained on DTU at low resolution and are evaluated with 5 input views at a resolution of $832 \times 1152$, with all other hyperparameters kept consistent.

\paragraph{Structure Analysis of BDF.} Tab.~\ref{tab:ablation} analyzes the structure of the proposed BDF module. Compared with the baseline (a), the one-way Mono$\rightarrow$Cost variants (b) and (c) improve MVS reconstruction, showing that monocular structural priors benefit cost volume refinement. Adapting the monocular decoder in (c) further improves the results over the frozen variant in (b). Conversely, Cost$\rightarrow$Mono in (d) substantially improves monocular depth estimation but provides only a small MVS gain, suggesting that multi-view evidence effectively refines monocular representations. Combining both directions, the full BDF in (g) improves Overall from 0.276 to 0.268 and monocular rel from 2.15 to 0.92 over the strongest one-way variant (c), yielding the best joint result. Without monocular supervision (f), MVS remains competitive, but monocular rel deteriorates to 8.09. Removing monocular pretraining (e) degrades both branches. These results suggest that pretraining supplies the structural prior, while stage-wise MDE supervision helps maintain informative monocular representations during bidirectional fusion.

\paragraph{Plug-and-play Compatibility.}
Tab.~\ref{tab:plug} evaluates the compatibility of BDF as a plug-in module for different cascaded MVS frameworks. Integrating BDF into MVSMamba~\cite{mvsmamba} and MonoMVSNet~\cite{monomvsnet} consistently improves both MVS and MDE metrics, demonstrating that BDF is not limited to a specific baseline. Notably, the runtime increase in MVSMamba mainly stems from introducing the additional MDE branch, whereas BDF adds only 0.07 s and 260 MB to MonoMVSNet, which already incorporates MDE. Since BDF operates on the regularized cost volume and DPT feature, it can be conveniently inserted into existing cascaded MVS frameworks to enhance both MVS and MDE.

\paragraph{Comparison under Different ViT Variants.}
Tab.~\ref{tab:vit} compares BridgeMVS with MonoMVSNet under different monocular ViT variants. BridgeMVS consistently outperforms MonoMVSNet with ViT-S, ViT-B, and ViT-L, showing that BDF is effective across different monocular foundation backbones. Notably, these improvements are achieved without additional runtime overhead. Across different ViT variants, BridgeMVS maintains comparable or lower memory consumption and inference time than MonoMVSNet, demonstrating that BDF is an effective and efficient way to exploit monocular foundation models for MVS reconstruction.

\begin{table}[!t]
    \centering
    \small
    \caption{Comparison with MonoMVSNet~\cite{monomvsnet} using different ViT variants.}
    \begin{tabular}{lccccccc}
    \toprule
    Methods & \multicolumn{1}{c}{Overall$\downarrow$} & Acc.$\downarrow$ & \multicolumn{1}{c}
    {Comp. $\downarrow$} & \multicolumn{1}{c}{MAE $\downarrow$} & \multicolumn{1}{c}{RMSE $\downarrow$} & Mem. (MB) $\downarrow$ & Time (s) $\downarrow$ \\
    \midrule
    MonoMVSNet-S & 0.278 & 0.313 & \textbf{0.243} & 12.94 & 34.65 & 2822 & \textbf{0.24} \\
    \grayrow BridgeMVS-S & \textbf{0.274} & \textbf{0.299} & 0.249 & \textbf{7.21} & \textbf{21.90} & \textbf{2734} & \textbf{0.24} \\
    \midrule
    MonoMVSNet-B & 0.278 & \textbf{0.294} & 0.262 & 10.91 & 30.52 & 3354 & 0.35 \\
    \grayrow BridgeMVS-B & \textbf{0.268} & 0.297 & \textbf{0.239} & \textbf{6.56} & \textbf{20.71} & \textbf{3132} & \textbf{0.34} \\
    \midrule
    MonoMVSNet-L & 0.281 & 0.311 & 0.251 & 10.37 & 29.61 & 4490 & 0.65 \\
    \grayrow BridgeMVS-L & \textbf{0.266} & \textbf{0.295} & \textbf{0.237} & \textbf{6.32} & \textbf{20.01} & \textbf{4016} & \textbf{0.64} \\
    \bottomrule
    \end{tabular}
    \label{tab:vit}
\end{table}

\section{Conclusion}

In this paper, we propose BridgeMVS, a cascaded multi-stage MVS framework that bidirectionally fuses monocular depth features and multi-view cost volumes. The proposed BDF module serves as a plug-and-play component for existing cascaded MVS frameworks, where monocular structural priors refine MVS cost volumes and multi-view geometric evidence improves monocular representations. Extensive experiments show that BridgeMVS achieves state-of-the-art performance on multiple MVS benchmarks and also improves monocular depth estimation.

\section*{Acknowledgments}
This work was supported by the Beijing Natural Science Foundation (No. 4262058, No. L257003) and Fundamental Research Funds for the Central Universities (No. FRF-TP-25-033).

\bibliographystyle{IEEEtran}
\bibliography{reference}


\appendix

\section{More Experimental Results}
\label{sec:more_exp}

\subsection{Comparison with Feed-forward Reconstruction Models}

Tab.~\ref{tab:feedforward} compares BridgeMVS with recent feed-forward reconstruction models, including VGGT~\cite{vggt}, $\pi^3$~\cite{pi3}, MASt3R~\cite{mast3r}, and Depth Anything 3~\cite{da3}, using the official DTU~\cite{dtu} point cloud metrics. For our runs of VGGT, $\pi^3$, and Depth Anything 3, predicted depths are aligned to GT through least-squares fitting and fused using GT camera poses, following the VGGT evaluation protocol. ``Known GT Camera'' indicates whether GT cameras are used to derive depth or geometry before fusion, rather than whether GT poses are used during fusion. MASt3R obtains geometry by triangulating feature correspondences using GT cameras. BridgeMVS achieves the lowest listed Overall error of 0.268 in its calibrated MVS setting. These results demonstrate its effectiveness for calibrated reconstruction, while differences in camera assumptions and the GT-depth alignment used for feed-forward predictions make this comparison additional quantitative context rather than a strictly equivalent evaluation.

\begin{table}[ht]
    \centering
    \small
    \caption{Comparison with feed-forward reconstruction models of point clouds on DTU.}
    \label{tab:feedforward}
    \begin{tabular}{lcccc}
        \toprule
        Methods & \shortstack{Known GT Camera} & Overall $\downarrow$ & Accuracy $\downarrow$ & Completeness $\downarrow$ \\
        \midrule
        VGGT (paper)~\cite{vggt} & $\times$ & 0.382 & 0.389 & 0.374 \\
        VGGT (our run)~\cite{vggt} & $\times$ & 0.525 & 0.634 & 0.416 \\
        $\pi^3$ (our run)~\cite{pi3} & $\times$ & 0.921 & 1.025 & 0.817 \\
        Depth Anything 3 (our run)~\cite{da3} & $\times$ & 1.235 & 1.432 & 1.038 \\
        \midrule
        MASt3R (paper)~\cite{mast3r} & \checkmark & 0.374 & 0.403 & 0.344 \\
        Depth Anything 3 (our run)~\cite{da3} & \checkmark & 1.210 & 1.410 & 1.010 \\
        \grayrow BridgeMVS (Ours) & \checkmark & \textbf{0.268} & \textbf{0.297} & \textbf{0.239} \\
        \bottomrule
    \end{tabular}
\end{table}

\subsection{Comparison with Alternative Fusion Strategies}
\label{sec:fusion_comparison}

We compare BDF with static bidirectional fusion and BridgeDepth-style Iterative Bidirectional Alignment (IBA) in Tab.~\ref{tab:fusion_comparison}. All variants use the same BridgeMVS backbone, training schedule, and DTU evaluation setting (5 views at $832 \times 1152$ resolution).

\begin{table*}[ht]
    \centering
    \small
    \setlength{\tabcolsep}{2.5pt}
    \caption{Comparison of alternative fusion strategies on DTU. MDE metrics are reported after GT alignment of the auxiliary output, which requires multi-view inputs. Params counts only the parameters of the fusion modules.}
    \label{tab:fusion_comparison}
    \begin{tabular}{l*{10}{c}}
        \toprule
        \multirow{2}{*}{Fusion Strategies} & \multicolumn{5}{c}{MVS branch} & \multicolumn{2}{c}{MDE branch} & \multicolumn{3}{c}{Model Efficiency} \\
        \cmidrule(lr){2-6}\cmidrule(lr){7-8}\cmidrule(lr){9-11}
        & Overall $\downarrow$ & Acc.\ $\downarrow$ & Comp.\ $\downarrow$ & MAE $\downarrow$ & RMSE $\downarrow$ & $\mathrm{rel}$ $\downarrow$ & $\tau$ $\uparrow$ & Time $\downarrow$ & Mem. $\downarrow$ & Params $\downarrow$ \\
        \midrule
        Concat, $C \times D$ & 0.278 & 0.289 & 0.268 & 6.78 & 21.32 & 2.03 & 82.1 & \textbf{0.27} & 3146 & 1.06 \\
        Add, $C \times D$ & 0.283 & \textbf{0.288} & 0.278 & 6.81 & 21.23 & 2.07 & 81.4 & \textbf{0.27} & 2974 & 1.03 \\
        Concat, per-depth & 0.277 & 0.293 & 0.260 & 7.01 & 22.00 & 2.04 & 82.1 & \textbf{0.27} & 3104 & 0.87 \\
        Add, per-depth & 0.279 & 0.300 & 0.257 & 7.15 & 22.48 & 2.06 & 82.1 & \textbf{0.27} & \textbf{2904} & \textbf{0.85} \\
        \midrule
        IBA~\cite{bridgedepth} & 0.276 & 0.290 & 0.262 & 7.95 & 23.55 & 1.21 & 90.5 & 0.50 & 13968 & 0.90 \\
        \midrule
        \grayrow BDF (Ours) & \textbf{0.268} & 0.297 & \textbf{0.239} & \textbf{6.56} & \textbf{20.71} & \textbf{0.92} & \textbf{93.6} & 0.34 & 3132 & 0.94 \\
        \bottomrule
    \end{tabular}
\end{table*}

\paragraph{Static bidirectional fusion.}
We construct four static baselines by combining two cost-volume organizations with concatenation or element-wise addition. In the $C \times D$-folded variants, the cost-channel and depth-hypothesis dimensions are merged into a single feature dimension. The resulting cost volume and the monocular feature are projected to the same dimension before fusion. In the per-depth variants, the depth dimension is retained, and the projected monocular feature is fused with each projected cost hypothesis. Each shared fused representation generates updates for both branches without input-dependent affinity weights. BDF achieves best performance than all four variants with comparable memory usage and runtime. The improvements across both branches support the benefit of selectively incorporating cross-branch evidence through direction-specific adaptive aggregation.

\paragraph{BridgeDepth-style IBA.}
We replace BDF with a BridgeDepth-style IBA module using local bidirectional cross-attention. With similar fusion parameters, BDF improves Overall from 0.276 to 0.268, MVS depth MAE from 7.95 to 6.56, and auxiliary monocular relative error from 1.21 to 0.92. It also reduces runtime from 0.50\,s to 0.34\,s and memory usage from 13,968\,MB to 3,132\,MB. This comparison concerns the transplanted fusion mechanisms rather than the complete native pipelines: IBA performs alignment at a single post-DPT stage, whereas its adaptation here operates on full MVS depth volumes at all four cascade stages, substantially increasing memory requirements. BDF instead uses cross-branch affinities to guide aggregation of the target branch's own values, supporting multi-stage interaction without direct cross-branch value transfer over all depth hypotheses.

\section{More Visualization Results}
\label{sec:more_vis}

\begin{figure}[ht]
  \centering
   \includegraphics[width=1.0\textwidth]{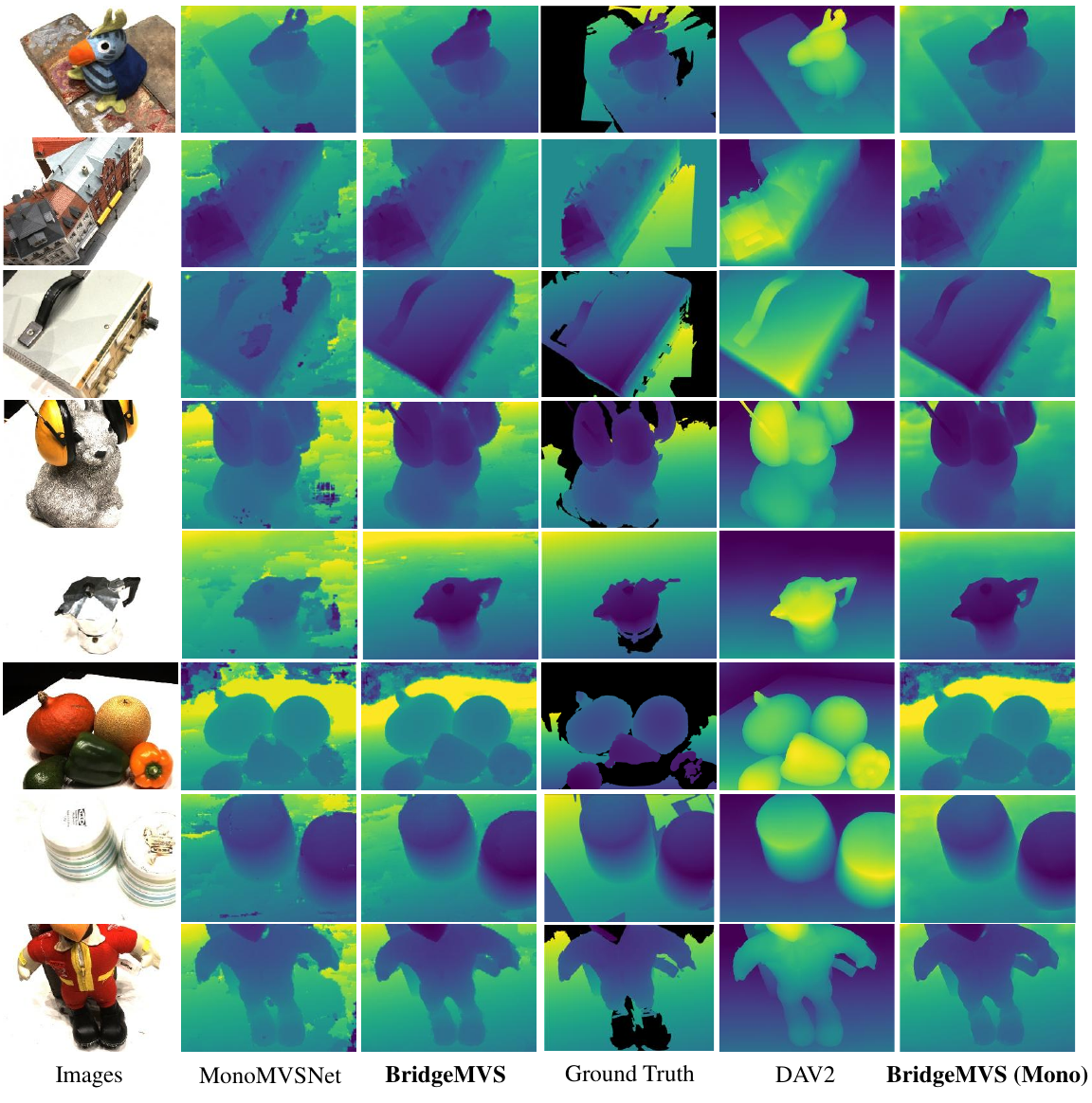}
   \caption{Additional depth maps visualization comparing the MonoMVSNet, BridgeMVS, Ground Truth, Depth Anything V2 (DAV2), and BridgeMVS (Mono) on the DTU \cite{dtu} dataset.}
   \label{fig:bridgemvs_dtu_depth_all}
\end{figure}

\begin{figure}[ht]
  \centering
  \includegraphics[width=1.0\linewidth]{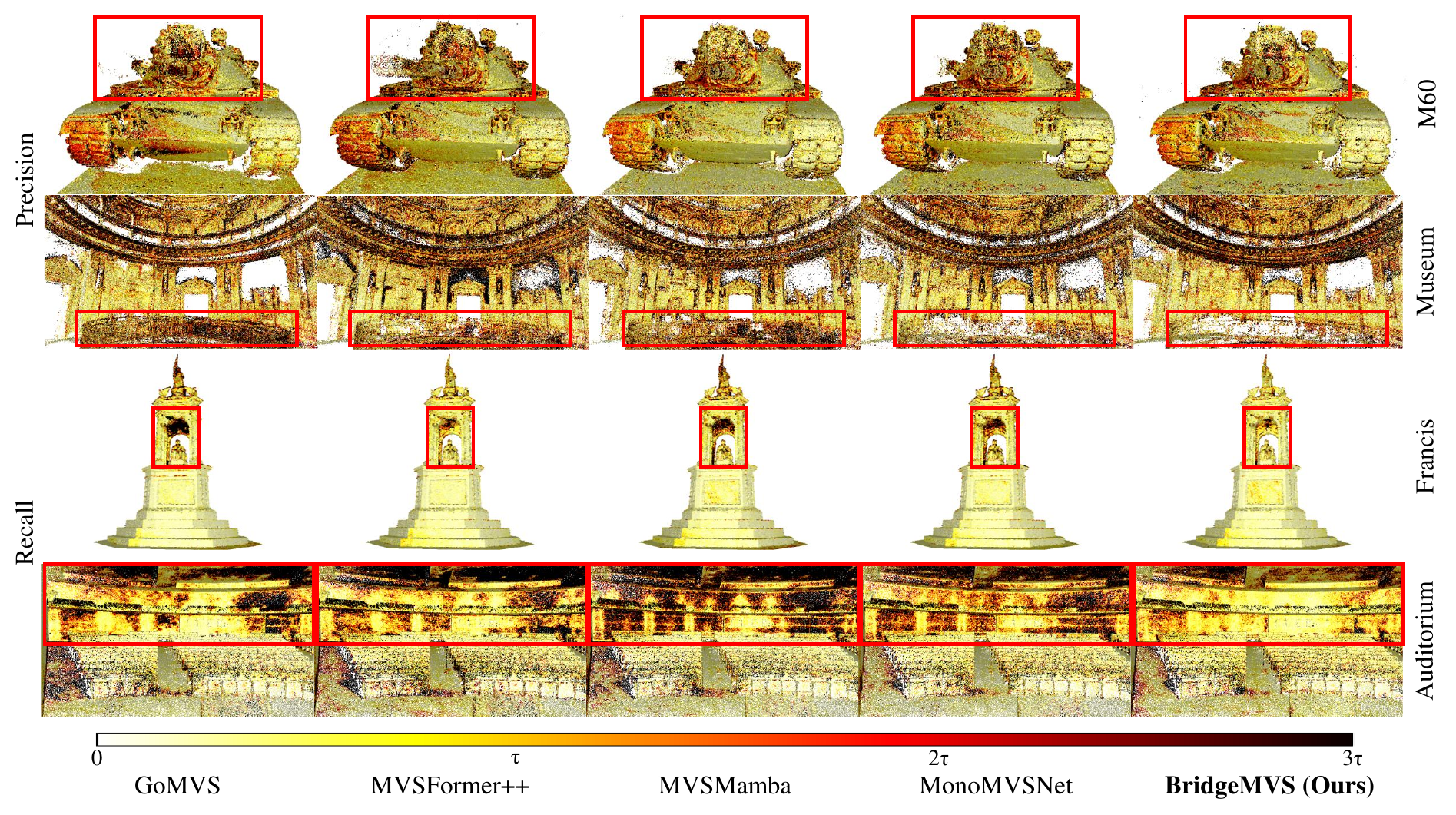}
  \caption{Qualitative comparison of reconstructed point clouds on the Tanks-and-Temples benchmark. Brighter regions indicate lower reconstruction errors under the corresponding distance threshold $\tau$.}
  \label{fig:bridgemvs_tnt}
\end{figure}

\begin{figure}[ht]
  \centering
  \includegraphics[width=1.0\linewidth]{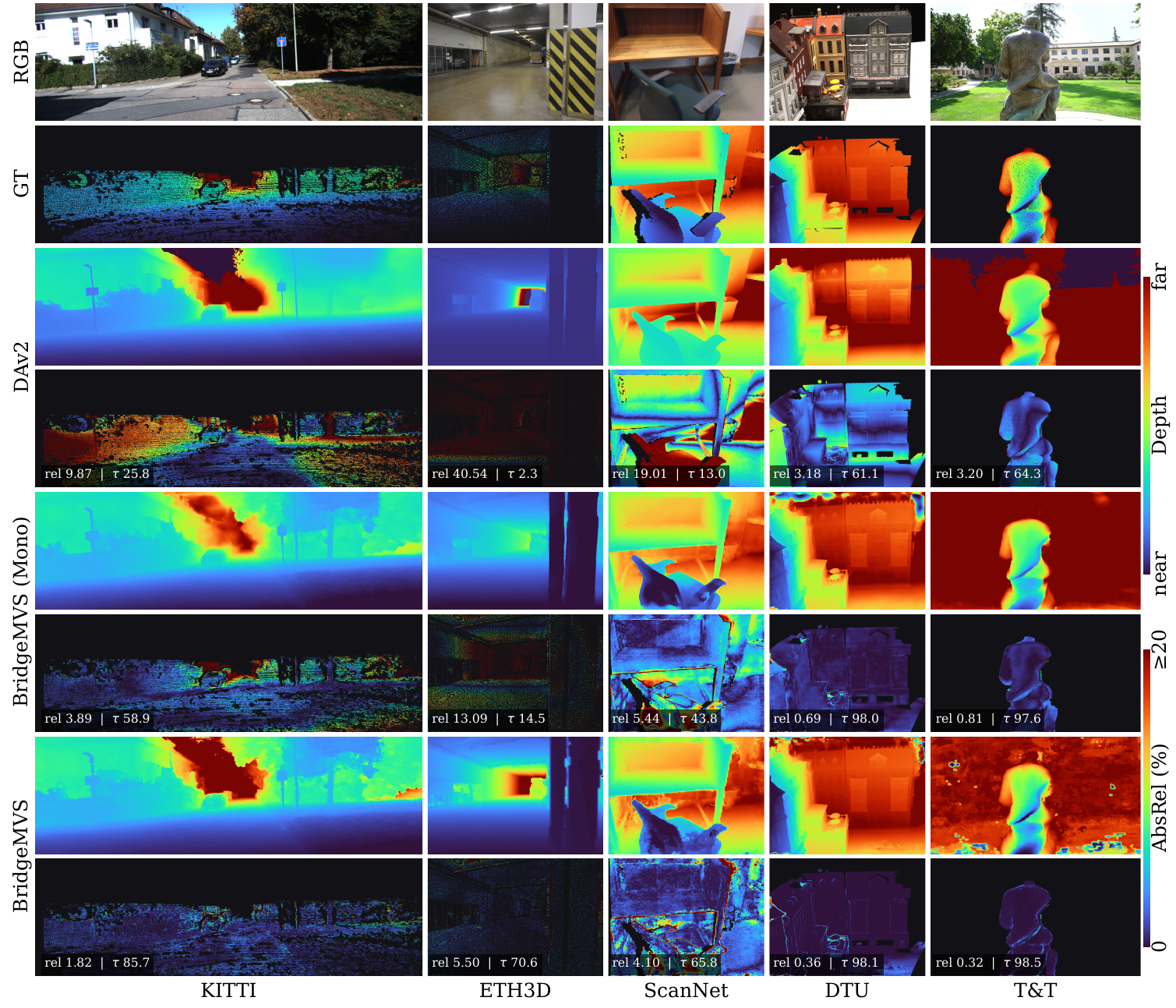}
  \caption{Qualitative comparison of depth maps and absolute relative error maps on the RobustMVD~\cite{rmvd} benchmark. For each method, the first row shows the predicted depth maps and the second row shows the corresponding error maps.}
  \label{fig:bridgemvs_rmvd}
\end{figure}

Fig.~\ref{fig:bridgemvs_dtu_depth_all} presents additional depth map comparisons with MonoMVSNet~\cite{monomvsnet} and Depth Anything V2~\cite{dav2} on the DTU~\cite{dtu} dataset. Benefiting from the bidirectional dynamic fusion between the MVS and MDE branches, BridgeMVS and BridgeMVS (Mono) mutually enhance each other and produce more accurate multi-view and monocular depth predictions. Fig.~\ref{fig:bridgemvs_tnt} further shows that our method achieves better accuracy and completeness in challenging regions on the Tanks-and-Temples benchmark~\cite{tanks}. Fig.~\ref{fig:bridgemvs_rmvd} shows qualitative results on RobustMVD~\cite{rmvd}, where BridgeMVS (Mono) benefits from multi-view geometric evidence provided by the MVS branch and produces more accurate monocular depth maps with lower relative errors.

\section{Limitations}
\label{limitations}

BridgeMVS remains subject to limitations of both monocular prediction and calibrated MVS. First, because the Cost-to-Mono update relies on the regularized cost volume, the enhanced monocular output requires multi-view inputs and is not a standalone single-image depth estimator. Second, BDF relies on complementary MDE and MVS cues for mutual enhancement. When both are unreliable, as can occur with complex materials or depth boundaries, bidirectional updates may propagate errors because BDF does not explicitly estimate cue reliability. Third, BridgeMVS assumes known intrinsics/extrinsics and an initial metric depth range, as required by its calibrated plane-sweep MVS backbone. Relative monocular depth alone cannot provide metric bounds. Future work will investigate uncertainty-aware fusion, robustness to perturbations in the required camera parameters, and using monocular depth priors to anchor the initial MVS depth range.

\clearpage



\end{document}